\documentclass[runningheads]{llncs}

\usepackage{eccv}

\usepackage{eccvabbrv}

\usepackage{graphicx}
\usepackage{booktabs}
\usepackage{amsmath}
\usepackage{siunitx}
\usepackage{float}

\usepackage[accsupp]{axessibility}  

\usepackage{bbm}

\usepackage{hyperref}

\usepackage{orcidlink}

\begin{document}

\title{The Nearest Target Is the Wrong One: Target Separation in Arc2Face Identity Unlearning}

\titlerunning{Target Separation in Arc2Face Identity Unlearning}

\author{Zeynel Tok\orcidlink{0009-0002-7209-4555}}

\authorrunning{Z.~Tok.}

\institute{University of Oxford, UK \\
\email{zeynel.tok@cs.ox.ac.uk}}

\maketitle

\begin{abstract}
Unlearning an identity from a face-conditioned generator by redirecting its conditioning embedding can silently fail if the redirected output is still verified as the original person. We show that this failure depends on a controllable choice of how far the redirection target lies from the forget identity in recognition space, and that the most intuitive target, the nearest neighbour, is the one most likely to cause it. We audit Arc2Face with a locked ArcFace protocol and a projection adapter that redirects identity conditioning before generation. On a hard-neighbour stress test built from the hardest \SI{0.5}{\percent} of eligible identities, four target-selection policies show a monotonic response: clean forgetting rises from 9/30 groups under the nearest hard target to 30/30 under the least similar one. Mean forget-identity re-identification falls from 51.9 to 0.0 while mean retention stays flat. This reflects successful redirection rather than outputs becoming unverifiable: 710 of 720 least-sim-hard generations arrive at the chosen target, with no leakage to unrelated identities. Re-verifying identical images with an independent recogniser (AdaFace) preserves that trend, correlating at $r=0.94$, arguing against a verifier artefact. Target separation is thus a first-order, reportable design variable for identity unlearning.


\end{abstract}

\section{Introduction}
\label{sec:intro}

Generative face models conditioned on identity embeddings can reproduce a specific person's likeness from only a handful of reference images \cite{photomaker,instantid}. This is directly privacy-relevant: a credible unlearning method must be evaluated against a protocol that would actually catch residual leakage, not just whether output pixels changed. Many unlearning studies rely on visual or prompt-level change as evidence of removal, but identity-conditioned generation demands a stricter test: a redirection target that is numerically different from the original identity can still be verifiably close enough to leak, so generated faces must be checked against a fixed recogniser and held-out references, not judged by eye.

\begin{figure}[t]
    \centering
    \includegraphics[width=0.9\textwidth]{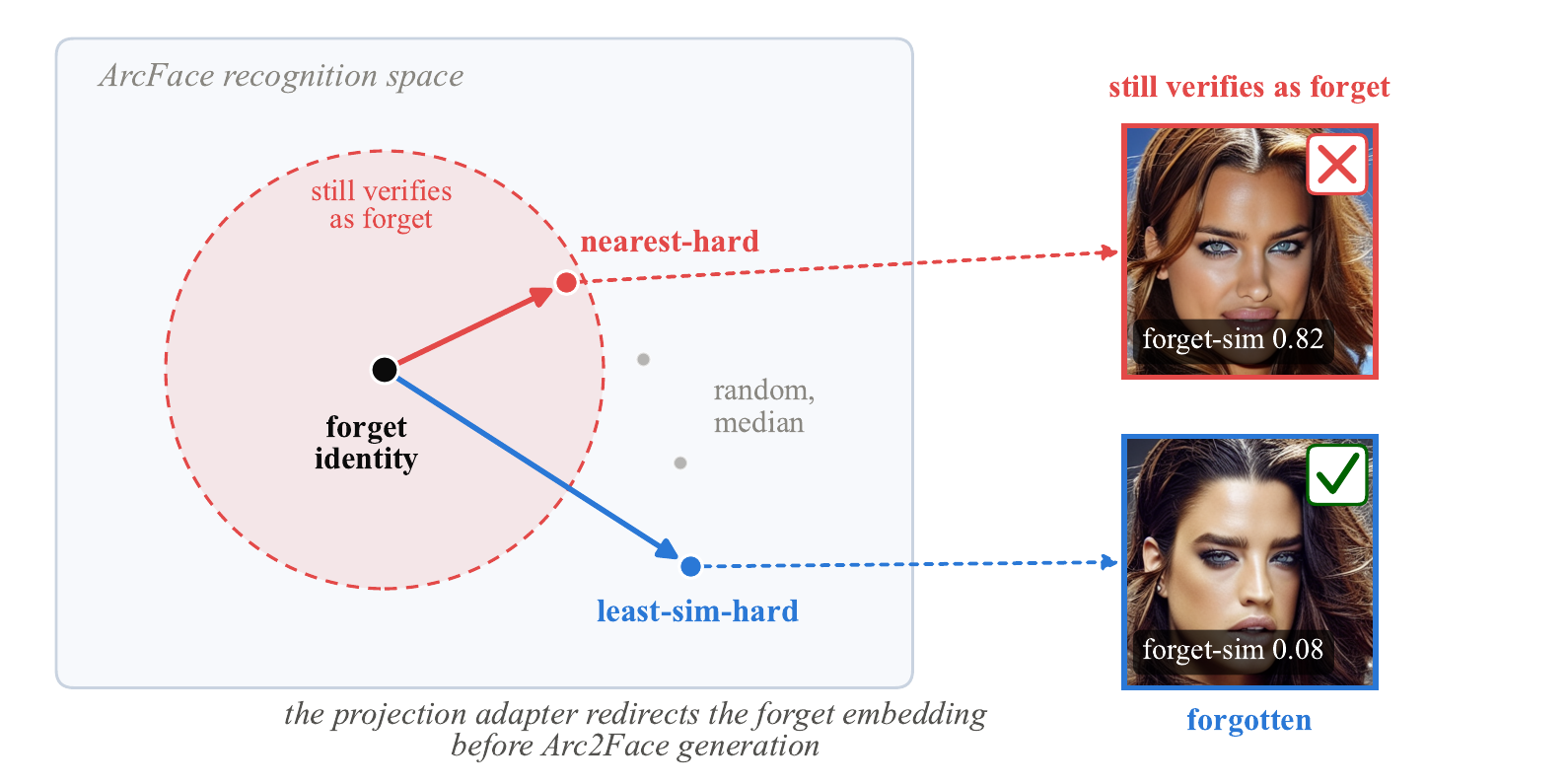}
    \caption{\textbf{Target separation decides whether identity unlearning forgets.}
    The projection adapter redirects the forget identity's ArcFace embedding before Arc2Face generation. Redirecting toward the nearest hard neighbour (nearest-hard) keeps the embedding inside the region that still verifies as the forget identity, so the generated face still verifies as the original person; redirecting toward the least similar hard neighbour (least-sim-hard) leaves that region and the face is no longer recognised. Faces are real generations for one forget identity under a matched reference and seed.}
    \label{fig:teaser}
\end{figure}

Using Arc2Face \cite{arc2face}, an identity-conditioned diffusion generator built on Stable Diffusion \cite{rombach2022ldm}, we ask whether the geometry of a redirection target, not merely its presence, determines success under hard-neighbour evaluation. A downstream cross-attention edit following a PIU style \cite{piu} identity replacement objective did not converge to a usable forget/retain operating point in our reproduction. To isolate the effect of redirection geometry more directly, we instead intervene directly in Arc2Face's identity conditioning pathway using the compact low-rank projection adapter defined in Eq.\eqref{eq:2}. The adapter redirects the forget identity's ArcFace \cite{arcface} embedding toward a selected target identity before generation while constraining retained identities to remain close to their original embeddings. We evaluate this intervention on a hard-neighbour stress test in which each forget identity is paired with its three nearest identities in ArcFace space, which serve as candidate redirection targets. Redirecting toward the nearest candidate often leaves generated images verifiable as the original identity. In contrast, selecting the least similar of these same three candidates increases source-target separation and resolves this failure mode (\Cref{fig:teaser}). 

Our contributions are:

\begin{itemize}
    \item A compact identity-pathway projection adapter used as a controlled intervention to isolate the effect of redirection geometry while leaving Arc2Face's generator weights unchanged.
    \item A four-policy hard-neighbour experiment showing that forget accuracy declines monotonically as the selected redirection target becomes less similar to the forget identity in ArcFace space while aggregate retention remains nearly unchanged.
\end{itemize}

\section{Related Work}
\label{sec:relatedwork}

\textbf{Identity-Conditioned Face Generation.}
Arc2Face \cite{arc2face} is an ID-conditioned face foundation model, generating diverse, photorealistic images from an ArcFace embedding alone rather than from a text prompt or ID embeddings in previous work \cite{instantid, photomaker}, making it a meaningful case study for identity unlearning. This capability has prompted a defensive line of work that aims to prevent harmful use of personal identities \cite{antidream}. Our setting is a complementary problem of removing an identity the generator can already reproduce. ArcFace \cite{arcface} is the face-recognition backbone we both condition on and verify against.

\textbf{Identity Unlearning and Replacement.}
Machine unlearning was formalised and later surveyed broadly \cite{cao, bourtoule, nguyen}. A complementary line of work studies how to evaluate and verify unlearning rather than how to perform it. These studies find that apparent forgetting is hard to certify as auditing can fail to detect residual information, so a target that evades detection has not necessarily been forgotten \cite{statistical, verification}. Our closed set verification result is an instance of the same pitfall. A forget identity can evade the group gallery match while its residual recognition similarity stays above the verification threshold (\Cref{sec:targetsep}). Studies framing unlearning in ID-conditioned diffusion models as identity replacement such as PIU \cite{piu} reassign the forget identity's conditioning embedding toward a proximity-selected anchor identity, combined with localised fine-tuning of identity-sensitive UNet cross-attention layers. 

While PIU reports positive results on Arc2Face, at the time of our experiments no official implementation was available and a paper faithful reproduction following its published hyperparameters did not converge to a usable forget/retain point in our codebase. This may reflect our reimplementation or unspecified details of the original setup rather than the method itself, so we do not report it as a comparative baseline. Instead, we adopt PIU's identity replacement framing but intervene upstream in the identity-conditioning pathway rather than fine-tuning cross-attention layers.

\textbf{Concept Erasure and Unlearning in Diffusion Models.}
A growing line of work removes semantic concepts from text-to-image diffusion models without full retraining, using a range of fine-tuning, closed-form model-editing, and continual-learning approaches
\cite{gandikota2023erasing,ablatingconcepts,uce,selectiveamnesia}. These methods target semantic concepts such as objects, styles, and unsafe content, and are typically verified by text-image similarity or human inspection. Our setting instead targets a specific person's identity and verifies removal against a face recogniser external to the image generator. A parallel line of work stress tests these methods and finds that removal is often superficial where supposedly erased concepts including identities can be recovered through adversarial prompts or concept inversion without changes to model weights \cite{circumventing}, establishing that erasure metrics can offer a false sense of security.

\textbf{Boundary Unlearning.}
Boundary Unlearning \cite{boundary} reframes class-level unlearning as a decision-boundary shift, showing that the destination to which forgotten samples are redirected matters for whether the shift succeeds. Its central observation has a direct analogue in our nearest-hard versus least-sim-hard comparison (Section~\ref{sec:results}): an intervention can move an embedding toward a chosen destination without that movement registering as forgetting, precisely because the destination is not sufficiently separated from the source.

\section{Preliminaries}
\label{sec:prelim}

\subsection{Problem Setup and Identity Verification}
\label{sec:problemsetup}

Let $\mathcal{I}$ denote the space of face images and $\phi:\mathcal{I}\to\mathbb{S}^{511}\subset\mathbb{R}^{512}$ the ArcFace embedding function \cite{arcface}, realised by the InsightFace \texttt{antelopev2} recognition model, mapping a detected face crop to a 512-dimensional, $\ell_2$-normalised embedding; cosine similarity thus reduces to the inner product $\mathrm{sim}(a,b)=\langle\phi(a),\phi(b)\rangle$. Arc2Face \cite{arc2face} is a diffusion model conditioned on $\phi(a)$ for a single reference image $a$ rather than on a text prompt. This conditioning pathway is therefore a natural intervention point for identity unlearning.

For each benchmark group we work with seven identities: the forget identity $y_F$ and six retain identities $\mathcal{Y}_R$, the latter split into three hard-retain neighbours and three random-retain identities. Every identity has two disjoint sets of three real reference images: a fitting/conditioning set $R_y^{\mathrm{fit}}$, used to condition generation and train the adapter, and a held out evaluation set $R_y^{\mathrm{eval}}$ used only to build the verification gallery. Identity $y$'s gallery centroid $c_y$ is the $\ell_2$-normalised mean of its evaluation-reference embeddings. Separately, the adapter's training objective (Section~\ref{sec:adapter}) and target-selection policies (Section~\ref{sec:targetpolicies}) draw on a background pool $\mathcal{P}$ of all eligible CelebA images that are not evaluation references of any benchmark identity. $y$'s pool centroid $\bar{c}_y$ is defined exactly as $c_y$ but over its images in $\mathcal{P}$, so it never touches the evaluation references. This makes the benchmark split-faithful where fit references drive adapter training and generation, evaluation references define the gallery, and no evaluation reference enters training, target selection, or threshold calibration.

To verify a generated image $g$, we detect its largest face and embed it as $\phi(g)$. We score it against each identity in the group gallery, $\mathcal{Y}_{\mathrm{grp}}=\{y_F\}\cup\mathcal{Y}_R$, by cosine similarity to that identity's gallery centroid, $s_y(g)=\langle\phi(g),c_y\rangle$ (an inner product, since all embeddings are unit-normalised), and take the closest gallery identity $\hat{y}(g)=\arg\max_{z\in\mathcal{Y}_{\mathrm{grp}}}s_z(g)$ as the closed-set prediction. We count $g$ as a verified match to $y$ only when $y$ is that prediction \emph{and} its score is at least the fixed threshold $\tau$:
\begin{equation}
\mathrm{match}(g,y)=\mathbbm{1}\big[\hat{y}(g)=y \ \wedge\ s_y(g)\ge\tau\big].
\end{equation}
This joint rule combines threshold verification with closed-set identification: a generated face can clear $\tau$ for
$y_F$ yet still fail to match it if a retain identity is closer. Detector based processing of the fitting/conditioning references uses $\mathrm{det\_thresh}=0.5$ during target selection, adapter training, and generation. The held-out evaluation pipeline uses $\mathrm{det\_thresh}=0.1$; a conventional default of $0.5$ here substantially reduced detected-face coverage in preliminary tests, producing metrics dominated by detector dropout rather than unlearning behaviour. Images with no detected face are excluded before metric computation rather than counted as non-matches, so all FA and RA calculations refer to detected faces, not attempted generations. Detection coverage is high and essentially constant across policies, so this exclusion does not advantage any policy's forget accuracy.

\subsection{Threshold Calibration}
\label{sec:thresholdcalibration}

We calibrate the verification threshold $\tau$ once, before evaluating any unlearning method, using a real-image calibration pool with all evaluation references removed. Following standard practice, we choose $\tau$ from different-identity (impostor) pairs so that the empirical false-accept rate (FAR) does not exceed $\alpha=0.01$. With $23{,}856$ impostor pairs, the closest attainable rate below this target accepts $238$ pairs, giving $\widehat{\mathrm{FAR}}(\tau)=238/23{,}856=0.009977$ as a 239th pair would exceed $\alpha=0.01$. The similarity score of this $238$th pair establishes our cutoff at $\tau=0.1465$, which is frozen for all evaluations.

\subsection{Metrics}
\label{sec:metrics}

Let $G_F$ denote evaluated generated images with the forget role, each conditioned on the forget identity $y_F$, and let $G_R$ denote evaluated generated images with the retain role, each carrying an intended identity $y(g)\in\mathcal{Y}_R$. We report the residual forget-identity re-identification rate as \textbf{Forget Accuracy} (FA, lower is better):
\[
\mathrm{FA}=\frac{100}{|G_F|}\sum_{g\in G_F}\mathrm{match}(g,y_F).
\]
\textbf{Retain Accuracy} (RA, higher is better) is
\[
\mathrm{RA}=\frac{100}{|G_R|}\sum_{g\in G_R}\mathrm{match}(g,y(g)).
\]

\textbf{Erasure-Retention Balance} (ERB) is computed per benchmark group as the harmonic mean of that group's own erasure accuracy,
$\mathrm{EA}=100-\mathrm{FA}$, and RA, with ERB set to zero if either term is zero. Reported ERB values are the mean of these per-group scores across the cohort, not the harmonic mean of the cohort's aggregate FA/RA. These two quantities generally differ when per-group FA or RA varies across the cohort, the gap is most visible when a group's FA reaches $100$ and therefore zero-floors that group's ERB regardless of its RA.

\textbf{IdentityLeak@8} measures worst-case leakage over generation seeds. For each forget identity and fit reference, we attempt eight fixed generation seeds. After excluding generations with no detected face, IdentityLeak@8 is the fraction of forget-reference blocks in which at least one detected output from those eight attempted seeds verifies as the forget identity. On groups with a hard/random retain split, we additionally report $\mathrm{RA}_{\mathrm{hard}}$ and $\mathrm{RA}_{\mathrm{random}}$, restricting RA
to the hard-retain and random-retain identities respectively, together with the hard-neighbour retention gap $\mathrm{RA}_{\mathrm{hard}}-\mathrm{RA}_{\mathrm{random}}$, which indicates whether retention cost concentrates on the hard neighbours nearest the forget identity. Because the forget identity is redirected onto one of the hard-retain identities, we also report $\mathrm{RA}_{\mathrm{target}}$, the same RA calculation restricted to retain-role generations whose intended identity is the selected redirection target.

\subsection{Benchmark Construction}
\label{sec:benchmarkconstruction}

We construct a deliberately adversarial stress test for local identity entanglement rather than an estimate of population average unlearning performance. We focus on the extreme hard-neighbour tail, where nearest identities are most confusable and therefore most likely to expose the nearest-neighbour redirection failure studied here. We do not extend the main benchmark to milder $1$--$2\%$ cohorts.

The benchmark is constructed from CelebA \cite{liu2015faceattributes}, restricted to identities with at least 15 eligible images. For each eligible identity, we compute its mean ArcFace cosine distance to its three nearest centroid neighbours. Identities with the smallest mean distances are the hardest cases because their nearest neighbours are closest in recognition space. 

The full eligible pool contains $5{,}964$ identities. For benchmark construction we exclude the 8 identities used in the provided public split from the GenMu 2026 challenge \cite{genmu2026}, leaving $5{,}956$ eligible non-public identities. We use a 30-identity cohort from the hardest end of this non-public pool, corresponding to the hardest $\SI{0.5}{\percent}$. Each benchmark group contains one forget identity, its three hard-retain neighbours, and three random-retain identities drawn under a fixed seed. With three fit references and eight generation seeds per identity, this gives $7\times3\times8=168$ generated images per group.

\begin{figure}[t]
    \centering
    \includegraphics[width=0.7\textwidth]{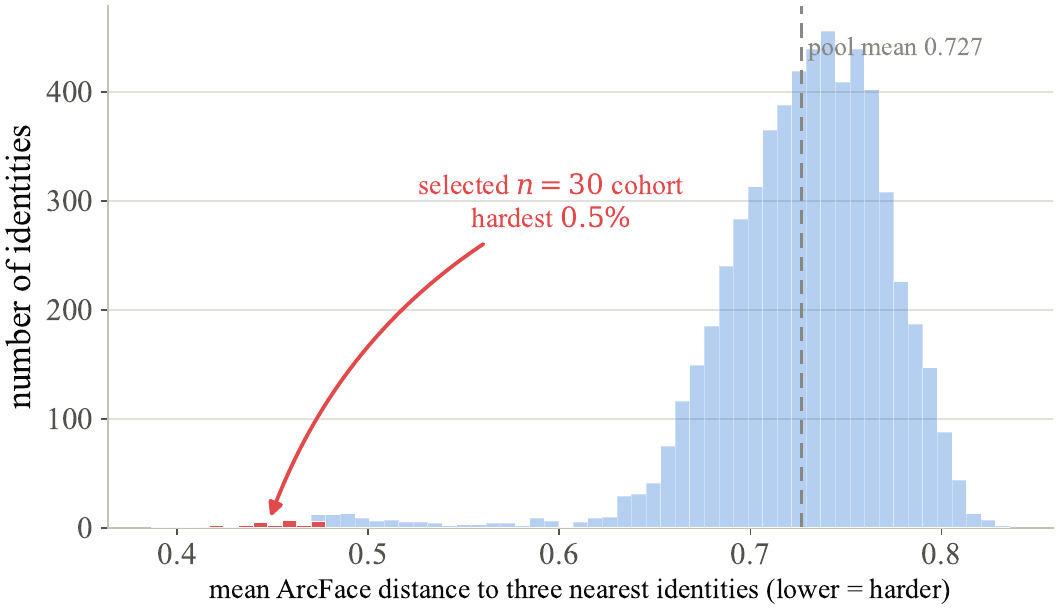}
    \caption{\textbf{The hard-neighbour cohort sits in the extreme tail.} Distribution of the identity-entanglement hardness score (mean ArcFace cosine distance to the three nearest identities) over the full eligible pool. The selected cohort lies in the low-distance tail (hardest $\SI{0.5}{\percent}$), where nearest neighbours are most confusable in recognition space.}
    \label{fig:hardness}
\end{figure}

The hardness-score distribution (\Cref{fig:hardness}) has mean $0.727$ and standard deviation $0.050$, with no sharp discontinuity separating the selected identities from the rest of the pool. The $10$th, $20$th, and $30$th hardest identities have $z$-scores of $-5.66$, $-5.33$, and $-5.02$, respectively, confirming that the selected cohort lies in the extreme low-distance tail rather than at an arbitrary cutoff. Section~\ref{sec:sensitivity} shows that the target-selection cohort size has stable results across the 10, 20, and 30-identity sized cohorts.

\section{Method}
\label{sec:method}

\subsection{Identity-Pathway Projection Adapter}
\label{sec:adapter}

We modify the identity embedding Arc2Face conditions on before it reaches the diffusion model. Given a normalised
ArcFace embedding $x=\phi(r)\in\mathbb{R}^{512}$, we learn a rank-$8$ residual
adapter
\begin{equation} \label{eq:2}
A_\theta(x)=\mathrm{normalize}\big(x+U(Dx)\big),
\end{equation}
with $D\in\mathbb{R}^{8\times512}$ and $U\in\mathbb{R}^{512\times8}$. This adds
$8{,}192$ learned parameters and modifies only the identity-conditioning
pathway, leaving the generator weights untouched.

The adapter is trained per forget identity with three cosine distance terms: forget ($\mathcal{L}_{\mathrm{forget}}$), retain ($\mathcal{L}_{\mathrm{retain}}$), and preserve ($\mathcal{L}_{\mathrm{preserve}}$), plus a regularisation term ($\mathcal{L}_{\mathrm{reg}}$). The forget loss $\mathcal{L}_{\mathrm{forget}}$ redirects the forget identity's fit reference embeddings toward the train pool centroid of a selected target identity. How far this redirection must go to register as forgetting is the subject of Section~\ref{sec:targetpolicies}. The retain loss $\mathcal{L}_{\mathrm{retain}}$ keeps retain identities' fit embeddings close to their originals, while the preserve loss $\mathcal{L}_{\mathrm{preserve}}$ applies the same constraint to a background batch of $512$ real embeddings resampled each step from the train pool.

Because this background batch excludes the forget identity's own images, the preserve term never opposes the forget term on the same identity. The regularisation loss $\mathcal{L}_{\mathrm{reg}}$ penalises the realised residual $U(Dx)$ on the embeddings the adapter actually sees, rather than a data-independent norm on $U$ and $D$. We weight the retain, preserve, and regularisation terms by $\lambda_r=5.0$, $\lambda_p=2.0$, and $\lambda_{\mathrm{reg}}=0.001$ respectively, and train for $1{,}000$ AdamW steps at learning rate $0.03$.

At inference the adapter is applied uniformly to every identity embedding, forget and retain alike, with no identity-specific gate; such a gate would be a hidden decision boundary untested outside the tuned identities. We therefore describe the method as identity-pathway redirection, not erasure of the base model's weights.

\subsection{Target-Selection Policies}
\label{sec:targetpolicies}

The training objective leaves one choice unspecified - which target identity the forget-reference embeddings should be redirected toward. The hard-neighbour benchmark provides three hard-retain candidates per group: the forget identity's three nearest centroid neighbours. We compare four policies for choosing among these candidates: \textbf{nearest-hard}, which selects the candidate with the highest cosine similarity to the forget identity's fit-reference centroid; \textbf{random-hard}, a seeded uniformly random choice; \textbf{median-hard}, the middle-ranked candidate; and \textbf{least-sim-hard}, the candidate with the lowest cosine similarity.

All four policies choose from the same three hard-retain candidates within each group. We therefore do not compare hard targets against easy targets; instead, we compare which member of the same hard candidate set is used for redirection, making target separation the controlled comparison. All target-selection similarities, including the ``Mean sim.'' column in \Cref{tab:expanded}, are computed between the forget identity's fit-reference centroid and each candidate identity's train-pool centroid $\bar{c}_y$ (Section~\ref{sec:problemsetup}), never using evaluation references $R_y^{\mathrm{eval}}$.

\section{Experimental Setup}
\label{sec:setup}

All generation uses the released Arc2Face model at fp16 precision, with 25 inference steps, guidance scale 3.0, and $512\times512$ output resolution. These generation settings are fixed across all target-selection policies. For each identity, we generate from three fit references and eight fixed seeds, resulting in $168$ attempted generations per group of Section~\ref{sec:benchmarkconstruction}.

The projection adapter is trained separately for each forget identity and target-selection policy. Unless stated otherwise, we select the checkpoint using the forget, retain, and preserve losses defined in Section~\ref{sec:adapter}, evaluated only on the training pool. Specifically, we choose the checkpoint with the highest $S_{\mathrm{ckpt}}$.
\[
S_{\mathrm{ckpt}}
=
(1-\mathcal{L}_{\mathrm{forget}})
+
0.5(1-\mathcal{L}_{\mathrm{retain}})
+
0.5(1-\mathcal{L}_{\mathrm{preserve}}),
\]
where no evaluation-reference image contributes to checkpoint selection.

Results below are reported on the complete canonical $n=30$ hard-neighbour benchmark unless stated otherwise (Section~\ref{sec:benchmarkconstruction}) across all four target-selection policies, including the $\mathrm{RA}_{\mathrm{target}}/\mathrm{RA}_{\mathrm{hard}}/\mathrm{RA}_{\mathrm{random}}$ retention split (Section~\ref{sec:retentionsplit}).

\section{Results}
\label{sec:results}

\subsection{Hard-neighbour Benchmark}
\label{sec:hardneighbour}

The hard-neighbour benchmark (Section~\ref{sec:benchmarkconstruction}) tests whether forgetting holds when the forget identity lies in a dense local ArcFace neighbourhood. \Cref{tab:expanded} reports all $30$ groups under all four target-selection policies. Nearest-hard targeting leaves substantial residual leakage: only 9/30 groups reach FA$=0$, and many redirected samples still verify as the forget identity. In contrast, least-sim-hard reaches FA$=0$ on all 30 groups while using the same benchmark, adapter architecture, training objective, and hard-retain candidate set. Mean RA is essentially unchanged across policies ($88.75$--$88.86$), so the large reduction in FA is not explained by broad degradation of retained-identity generation.

\begin{table}[t]
    \centering
    \caption{Hard-neighbour benchmark ($n=30$ cohort) across four target-selection policies. Lower FA and Leak@8 indicate better forgetting, higher RA and ERB indicate better retention/balance. Forgetting improves monotonically as mean target similarity decreases, while retention remains
nearly constant. Except for FA\(=0\) groups, all columns are group-wise means.}
    \label{tab:expanded}
    \begin{tabular}{lrrrrrr}
        \toprule
        Policy & FA=0 groups & FA$\downarrow$ & RA$\uparrow$ & ERB$\uparrow$ & Leak@8$\downarrow$ & Target sim. \\
        \midrule
        nearest-hard & 9/30 & 51.94 & 88.79 & 50.34 & 66.67 & 0.8689 \\
        random-hard & 20/30 & 25.83 & 88.86 & 71.81 & 31.11 & 0.5460 \\
        median-hard & 28/30 & 2.64 & 88.75 & 92.35 & 6.67 & 0.3403 \\
        least-sim-hard & 30/30 & 0.00 & 88.86 & 93.84 & 0.00 & 0.2874 \\
        \bottomrule
    \end{tabular}
\end{table}

\subsection{Retention Split by Retain Role.}
\label{sec:retentionsplit}

\Cref{tab:retentionsplit} decomposes RA into hard-retain and random-retain identities on the full $n=30$ cohort, and reports two training-free face-image-quality scores on the same generations: the released \textbf{SER-FIQ} \cite{serfiq} implementation, with $T=100$ stochastic passes, and \textbf{GraFIQs-B2} \cite{grafiqs}, both with their authors' released ArcFace backbones. Within each group, each detected generation is compared only with the six real references of its intended identity, and we report the mean of the 30 group-level $P(\text{generated}\geq\text{real})$ values, orienting both scores so that higher is better. The baseline contains 960 unique CelebA images, corresponding to 1,260 group memberships; $50\%$ denotes no ordering advantage.

\begin{table}[t]
\centering
\caption{Retention split by hard- versus random-retain role, and face-image-quality scores on the $n=30$ cohort. Aggregate, hard-retain, and random-retain RA remain nearly unchanged across policies, whereas target-identity RA increases substantially with target separation. Neither quality score differs materially across policies.}

\label{tab:retentionsplit}
\begin{tabular}{lrrrrrr}
\toprule
Policy & Mean RA & $\mathrm{RA}_{\mathrm{target}}$ & $\mathrm{RA}_{\mathrm{hard}}$ & $\mathrm{RA}_{\mathrm{random}}$ & SER-FIQ$_{\geq\mathrm{real}}\uparrow$ & GraFIQs$_{\geq\mathrm{real}}\uparrow$ \\
\midrule
nearest-hard & 88.79 & 53.89 & 79.81 & 97.85 & 70.28\% & 10.64\% \\
random-hard & 88.86 & 77.08 & 79.96 & 97.85 & 70.48\% & 10.77\% \\
median-hard & 88.75 & 92.22 & 79.78 & 97.80 & 69.62\% & 10.81\% \\
least-sim-hard & 88.86 & 94.28 & 79.90 & 97.89 & 69.07\% & 11.12\% \\
\bottomrule
\end{tabular}
\end{table}

Across all four policies, $\mathrm{RA}_{\mathrm{hard}}$ remains near $80$ while $\mathrm{RA}_{\mathrm{random}}$ remains near $98$. Thus, retention errors are concentrated among hard neighbours rather than random retains, but this hard/random gap is stable across target-selection policies. The selected target identity is the exception: $\mathrm{RA}_{\mathrm{target}}$ rises from $53.89$ under nearest-hard to $94.28$ under least-sim-hard. The nearest hard target is therefore both the weakest forgetting choice and the one whose own retain generations suffer most; aggregate RA stays flat because this target-specific swing is balanced by the remaining hard-retain identities and consistently high random-retain RA. 

Both quality columns are flat across policies so the $\mathrm{RA}_{\mathrm{target}}$ swing above does not extend to image quality. Generated images outrank real references in $69$--$71\%$ with SER-FIQ comparisons but only $10$--$12\%$ under GraFIQs.

\subsection{Effect of Target Separation}
\label{sec:targetsep}

\begin{figure}[t]
    \centering
    \includegraphics[width=0.85\textwidth]{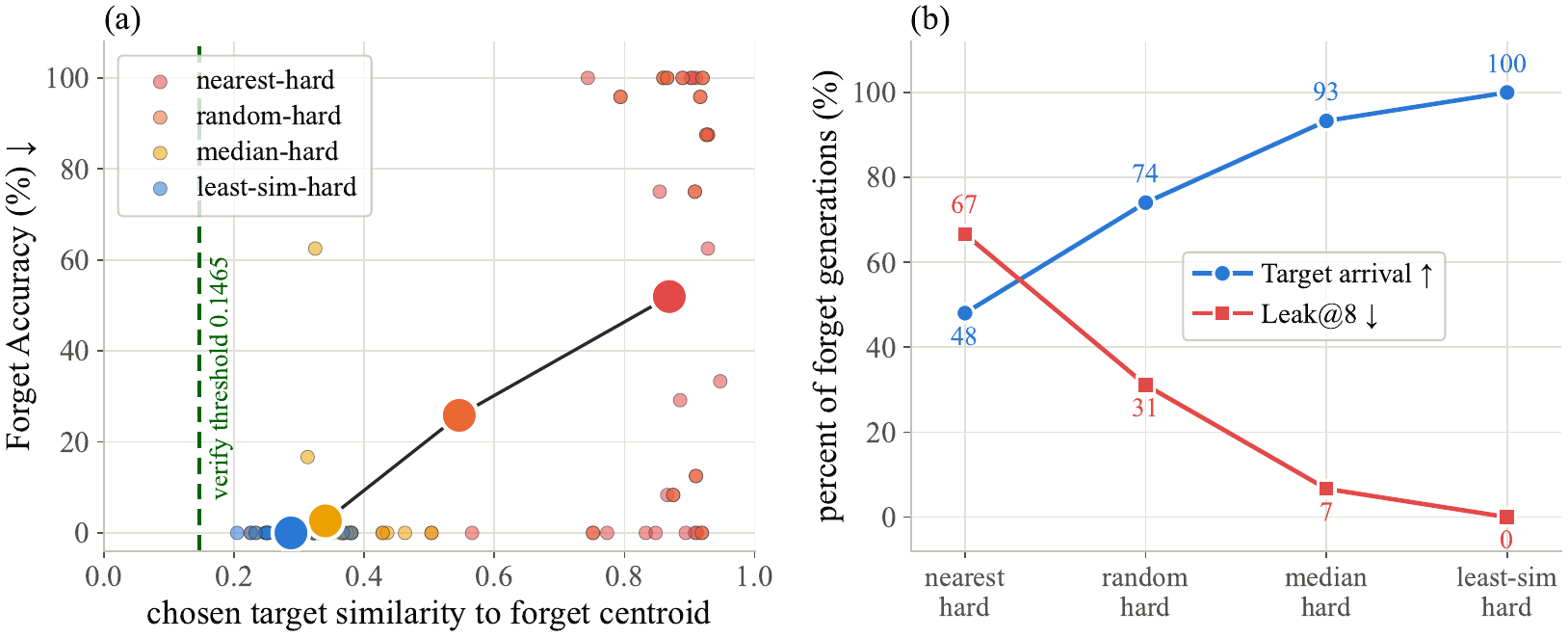}
    \caption{\textbf{Target separation drives forgetting through successful target redirection.} (a)~Per-group FA against chosen target similarity for all $120$ group-policy points with policy-mean anchors. FA falls as the chosen target moves away from the forget identity. (b)~As target separation increases, target arrival among detected forget generations rises from $48\%$ to $100\%$ while mean Leak@8 falls from $67\%$ to $0\%$.}
    \label{fig:mechanistic}
\end{figure}

\Cref{tab:expanded} shows a clear four-policy trend (from \Cref{fig:mechanistic}a): as the selected target moves farther from the forget identity in ArcFace space, residual forget verification falls. This is not a forgetting/retention trade-off, mean target similarity falls from $0.8689$ to $0.2874$ and mean FA from $51.94$ to $0.00$, while mean RA is mostly unchanged (\Cref{tab:expanded}). Random-hard and median-hard lie between the two endpoints in both target similarity and forget accuracy. Thus, the effect is not driven by a single nearest-hard/least-sim-hard comparison, but by a monotonic sweep over target separation.

We treat the target similarities as a controlled measure of relative separation within each hard-neighbour group, not as verification scores directly comparable to the pair-calibrated threshold $\tau$. The result suggests that nearest-hard fails because the redirection target remains too entangled with the forget identity: the adapter moves the conditioning embedding, but not far enough to prevent generated samples from re-verifying as the source. Least-sim-hard succeeds because it redirects toward a less entangled target while preserving retain identities.

\begin{figure}[t]
    \centering
    \includegraphics[width=0.7\textwidth]{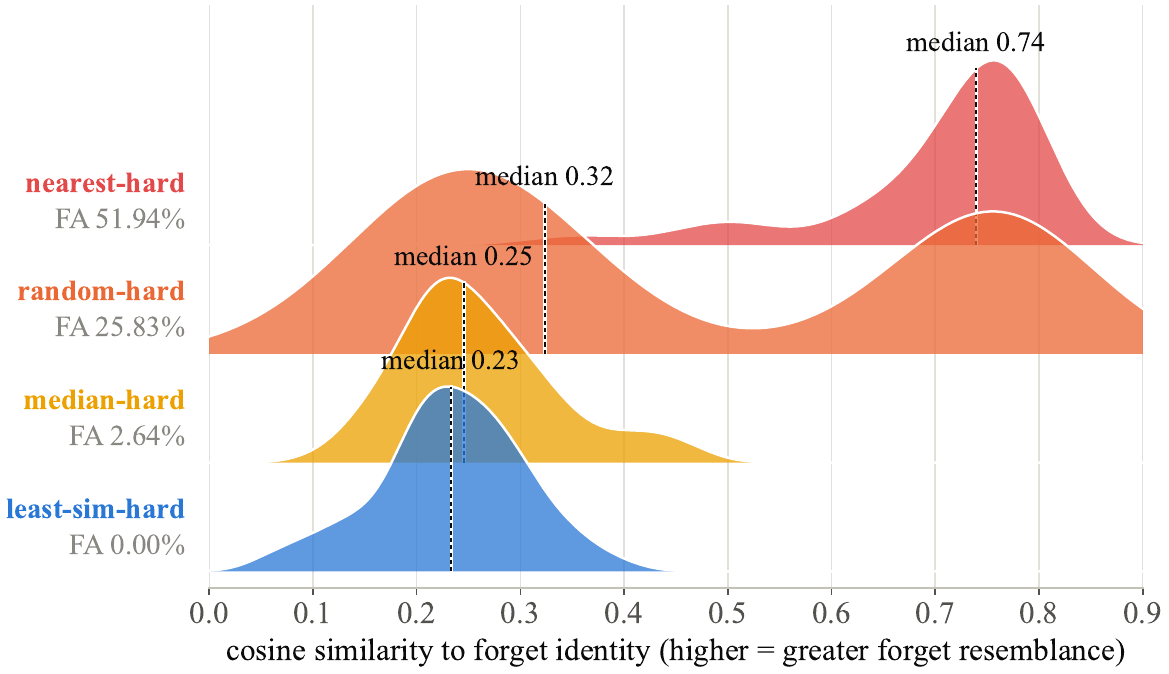}
    \caption{\textbf{Image-level forget resemblance collapses with target separation.} Per-policy distribution of the generated forget-role output's ArcFace cosine similarity to the forget identity, pooled across 30 identity groups. As target separation increases, median similarity falls from $0.74$ to $0.23$, and mean FA falls from $51.94\%$ to $0\%$.}
    \label{fig:forget}
\end{figure}

The low FA under least-sim-hard is associated with successful redirection to the selected target rather than merely making the outputs fail verification as the forget identity. Target classification among detected forget-role generations rises monotonically from $48.0\%$ under nearest-hard to $74.1\%$ under random-hard, $93.3\%$ under median-hard, and $100.0\%$ under least-sim-hard ($710/710$, \Cref{fig:mechanistic}b). Under the large-gallery check against the full eligible pool (Section~\ref{sec:limitations}), the least-sim-hard result also holds as thresholded verification: $100.0\%$ of detected forget-role generations verify as the selected target identity, with $0.0\%$ forget-role bystander leakage. Counting no-face attempts as non-arrivals still gives $710/720=98.6\%$ target arrival. The adapter therefore redirects the forget conditioning onto the selected, more separated identity rather than pushing it into an unverifiable region. The same shift is visible at the image level where the distribution of forget-role outputs' ArcFace similarity to the forget identity falls from a median of 0.74 under nearest-hard to 0.23 under least-sim-hard (\Cref{fig:forget}).

\subsection{Second-recogniser Robustness}
\label{sec:secondrecogniser}

ArcFace defines both the adapter's embedding space and the primary verification protocol, leaving open whether the results are specific to ArcFace geometry. We therefore re-verify every generated image from \Cref{tab:expanded}, the complete canonical $n=30$ cohort across all four policies, with no regeneration, using AdaFace \cite{adaface}. In our pipeline, AdaFace shares only face detection and 5-point alignment with ArcFace. The embedding network and FAR-\SI{1}{\percent}-calibrated verification threshold are separate. \Cref{tab:secondrecogniser} reports FA and RA for the identical generated images under both recognisers. The nearest-hard to least-sim-hard decline in FA reproduces under AdaFace, while RA remains flat for both recognisers ($88.6$--$88.9$). 

\begin{figure}[t]
    \centering
    \includegraphics[width=0.85\textwidth]{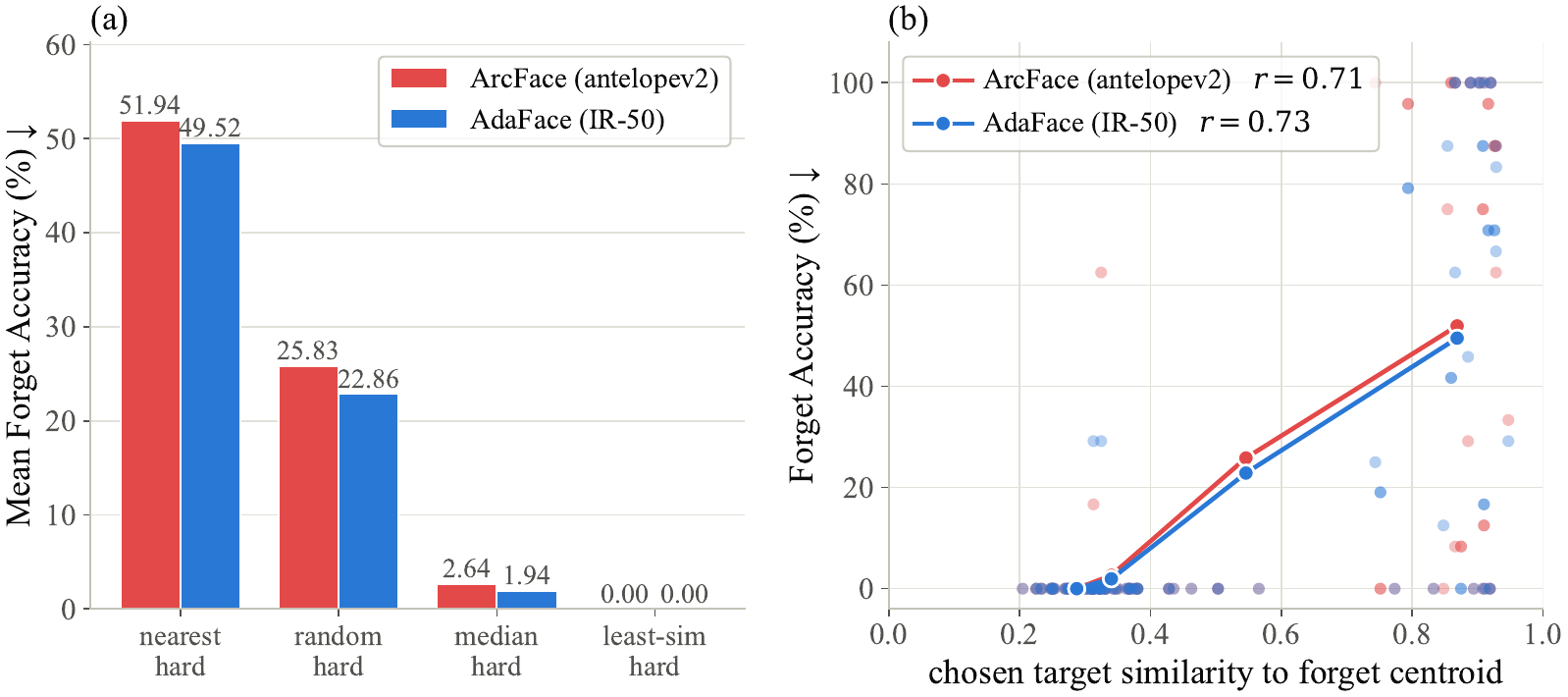}
    \caption{\textbf{The trend reproduces under a second recogniser.} (a) Mean FA per policy under ArcFace and AdaFace follows the same ordering. (b) The relationship between chosen target similarity and forget accuracy, using the same ArcFace-defined target-selection similarity as \Cref{fig:mechanistic}, reproduces under AdaFace.}
    \label{fig:secondrecogniser}
\end{figure}

\begin{table}[t]
\centering
\caption{Second-recogniser robustness on identical generated images from the
canonical $n=30$ cohort. Re-verification with AdaFace preserves the same
nearest-hard to least-sim-hard trend observed under ArcFace, with similar FA
and RA values.}
\label{tab:secondrecogniser}
\begin{tabular}{lrrrrrr}
\toprule
& \multicolumn{3}{c}{ArcFace} & \multicolumn{3}{c}{AdaFace} \\
Policy & FA=0 & Mean FA$\downarrow$ & Mean RA & FA=0 & Mean FA$\downarrow$ & Mean RA \\
\midrule
nearest-hard & 9/30 & 51.94 & 88.79 & 8/30 & 49.52 & 88.64 \\
random-hard & 20/30 & 25.83 & 88.86 & 20/30 & 22.86 & 88.64 \\
median-hard & 28/30 & 2.64 & 88.75 & 28/30 & 1.94 & 88.70 \\
least-sim-hard & 30/30 & 0.00 & 88.86 & 30/30 & 0.00 & 88.67 \\
\bottomrule
\end{tabular}
\end{table}
Using the same ArcFace-defined target-selection similarity as Section~\ref{sec:targetsep}, the target-similarity/FA trend is preserved under AdaFace: the pooled correlation is $r=0.71$ under ArcFace and $r=0.73$ under AdaFace. The recognisers' group-policy FA values also agree closely ($r=0.94$, mean absolute FA difference $4.6$ points), and both agree that FA$=0$ on 84/120 group-policy pairs. \Cref{fig:secondrecogniser} shows the per-policy comparison and overlapping policy curves. Since AdaFace uses a separate embedding network and a separately calibrated threshold, this result argues against the effect being only an ArcFace-verifier artefact. It does not establish representation-independent generalisation, because both the intervention and target-selection policies remain defined in ArcFace space.

\subsection{Benchmark-Size Sensitivity}
\label{sec:sensitivity}

To check that the ordering is not driven by one part of the cohort, we recompute all four policies on nested sub-cuts. On the
hardest-10 core, FA$=0$ groups rise from 5/10 (nearest-hard) to 10/10 on the remaining 3 policies. On the hardest-20, from 7/20 through 14/20 and 19/20 to 20/20; mean RA stays within about 1.5 percentage points of the full-cohort value in every experiment. As nested sub-cuts of one cohort, these are a sensitivity analysis, not independent replications.

\section{Discussion}
\label{sec:discussion}

Our projection adapter is identity-specific and redirects the conditioning embedding before generation, it is not a persistent edit to Arc2Face's base weights. Within that scope, however, the hard-neighbour sweep shows that target selection is not a cosmetic implementation detail. With the experimental procedure fixed, changing only the redirection target moves mean FA from $51.94$ to $0.00$ while RA remains essentially unchanged. The direction of this effect may appear predictable in hindsight, but the experiment establishes three less obvious properties of this relationship. First, prior identity replacement work favours proximity selected anchors, yet under the same groups, adapter, objective, and hard-candidate sets, the nearest target is precisely the one that leaves the most residual source verification, while the least similar target eliminates it. Second, the naive expectation is that target choice is absorbed by optimisation where the forget term explicitly pushes the conditioning embedding onto the chosen target, so a sufficiently trained adapter might reach any of the three candidates equally well. This expectation is false: under an identical objective and training budget, nearest-hard leaves mean FA at $51.94$ rather than falling toward zero. Third, greater separation does not reduce FA by degrading overall retention. Aggregate RA remains flat across the sweep, while the component that changes most, the redirection target's own retention ($\mathrm{RA}_{\mathrm{target}}$, \Cref{tab:retentionsplit}), improves from approximately $54$ to $94$ as separation increases. Thus, within this controlled setting, increased target separation improves forgetting without introducing a forgetting/retention trade-off.

The central implication is that identity-pathway unlearning should be evaluated as a local geometry problem, not only as a before/after removal problem. A redirection target can be numerically different from the forget identity yet remain close enough in the recognition neighbourhood for generated images to re-verify as the original identity. Conversely, increasing target separation within the same hard candidate set produces a graded reduction in residual forget verification. This makes target separation a measurable design variable and should be specified, reported, and stress-tested rather than treated as an incidental choice. This also affects how identity-unlearning benchmarks should be constructed. The hard-neighbour benchmark is designed for failure discovery rather than estimating how frequently this behaviour occurs across the full identity population. The present cohort asks whether redirection still succeeds when the available targets are deliberately confusable. We therefore do not claim that the observed effect size reported here represents average-case identity unlearning.

Several extensions follow directly. First, target selection should be optimised over a larger identity pool under explicit constraints on separation, retain risk, and bystander leakage, rather than restricted to benchmark-provided hard neighbours. Second, while Section~\ref{sec:retentionsplit} finds no detectable policy-dependent quality change under identity-matched official SER-FIQ and GraFIQs-B2, these scores do not establish preservation of non-identity content. Future evaluation should therefore measure perceptual and semantic fidelity directly, including pose, expression, attributes, and demographic presentation. Third, future work should test whether the same target-separation effect appears for persistent model edits, multiple forget identities, and identity-conditioned generators that do not use ArcFace as their conditioning representation.

\section{Limitations}
\label{sec:limitations}

Our study is intentionally narrow. The intervention and target-selection
policies are defined in ArcFace space because Arc2Face itself conditions on
ArcFace embeddings. The AdaFace re-verification in Section~\ref{sec:secondrecogniser} argues against the result being only an
ArcFace-verifier artefact, but it does not remove the dependence of the
intervention and target-selection rule on ArcFace geometry. The method is also adapter-based redirection rather than base-model erasure: it learns an identity-specific adapter applied before generation, and therefore does not
provide a persistent deletion guarantee for Arc2Face's weights.

The benchmark likewise studies a controlled but restricted setting. Each adapter is trained for one forget identity, and all four target-selection policies choose among three hard-retain candidates per group rather than searching over the full eligible identity pool. We also focus primarily on identity verification metrics. Section~\ref{sec:retentionsplit} reports two training-free face-image-quality scores and finds no material difference across policies, which weighs against a broad quality trade-off, but neither score is a calibrated, standalone quality benchmark, and we do not report perceptual, semantic-fidelity, or attribute-consistency checks, so we do not claim that successful redirection preserves non-identity content such as pose, expression, attributes, or demographic presentation.

The main verification protocol uses each group's seven identity gallery rather than a large-population search. As an additional check, we compare against all $5{,}964$ eligible identities: forget-role bystander leakage is $0.0\%$ across all four policies, while retain-role bystander leakage is $4.5$--$4.6\%$ and flat across policies. This suggests that nearest-hard failures are re-verifications of the forget identity rather than diffuse leakage to unrelated identities, but large-gallery verification remains a distinct operating regime rather than a replacement for the small-gallery protocol.

Our experiments use real hard-retain identities as redirection targets because they are part of the controlled benchmark. This should not be interpreted as a deployment recommendation. Redirecting a forget identity toward another real person may transfer privacy or misuse risk to the target identity. Practical systems should prefer synthetic, non-identifying, or explicitly consenting target representations, and should optimise target selection under constraints on separation, retain risk, and bystander leakage.

\section{Conclusion}
\label{sec:conclusion}

We presented a split-faithful audit of identity unlearning in Arc2Face. A compact projection adapter in the ArcFace conditioning pathway can suppress identity leakage, but the hard-neighbour benchmark shows that intervention placement alone is not enough. Within the hardest $\SI{0.5}{\percent}$ of the eligible non-public identity pool ($n=30$), least-sim-hard target selection improves clean-forgetting success from 9/30 groups to 30/30 relative to nearest-hard targeting while leaving retention nearly unchanged. This is stable across nested 10, 20, and 30-identity sub-cuts and reproduces under an independently trained second recogniser, arguing against a verifier-only ArcFace artefact. Within Arc2Face identity redirection, the result identifies target separation as a practical, measurable design axis that identity unlearning methods should report and stress test. Whether the same effect transfers to other conditioning representations remains an open question.

%
%
\bibliographystyle{splncs04}
\bibliography{main}

@String(CVPR  = {IEEE Conf. Comput. Vis. Pattern Recog.})

@String(ICCV  = {Int. Conf. Comput. Vis.})

@String(ECCV  = {Eur. Conf. Comput. Vis.})

@String(NeurIPS = {Adv. Neural Inform. Process. Syst.})

@String(CVPRW = {IEEE Conf. Comput. Vis. Pattern Recog. Worksh.})

@String(CVPR  = {CVPR})

@String(ICCV  = {ICCV})

@String(ECCV  = {ECCV})

@String(NeurIPS = {NeurIPS})

@String(CVPRW = {CVPRW})

@inproceedings{arc2face,
  title={Arc2face: A foundation model for id-consistent human faces},
  author={Papantoniou, Foivos Paraperas and Lattas, Alexandros and Moschoglou, Stylianos and Deng, Jiankang and Kainz, Bernhard and Zafeiriou, Stefanos},
  booktitle={European Conference on Computer Vision},
  pages={241--261},
  year={2024},
  organization={Springer}
}

@InProceedings{adaface,
    author    = {Kim, Minchul and Jain, Anil K. and Liu, Xiaoming},
    title     = {AdaFace: Quality Adaptive Margin for Face Recognition},
    booktitle = {Proceedings of the IEEE/CVF Conference on Computer Vision and Pattern Recognition (CVPR)},
    month     = {June},
    year      = {2022},
    pages     = {18750-18759}
}

@inproceedings{arcface,
  title        = {ArcFace: Additive Angular Margin Loss for Deep Face Recognition},
  author       = {Deng, Jiankang and Guo, Jia and Xue, Niannan and Zafeiriou, Stefanos},
  booktitle    = {IEEE/CVF Conference on Computer Vision and Pattern Recognition (CVPR)},
  pages        = {4690--4699},
  year         = {2019}
}

@inproceedings{boundary,
  title        = {Boundary Unlearning: Rapid Forgetting of Deep Networks via Shifting the Decision Boundary},
  author       = {Chen, Min and Gao, Weizhuo and Liu, Gaoyang and Peng, Kai and Wang, Chen},
  booktitle    = {IEEE/CVF Conference on Computer Vision and Pattern Recognition (CVPR)},
  pages        = {7766--7775},
  year         = {2023}
}

@inproceedings{liu2015faceattributes,
  title = {Deep Learning Face Attributes in the Wild},
  author = {Liu, Ziwei and Luo, Ping and Wang, Xiaogang and Tang, Xiaoou},
  booktitle = {Proceedings of International Conference on Computer Vision (ICCV)},
  month = {December},
  year = {2015} 
}

@inproceedings{gandikota2023erasing,
  title        = {Erasing Concepts from Diffusion Models},
  author       = {Gandikota, Rohit and Materzynska, Joanna and Fiotto-Kaufman, Jaden and Bau, David},
  booktitle    = {IEEE/CVF International Conference on Computer Vision (ICCV)},
  year         = {2023}
}

@misc{piu,
      title={PIU: Proximity-guided Identity Unlearning in ID-Conditioned Diffusion Models}, 
      author={Jose Edgar Hernandez Cancino Estrada and Mauro Díaz Lupone and Žiga Emeršič and Vitomir Štruc and Peter Peer and Darian Tomašević},
      year={2026},
      eprint={2605.22311},
      archivePrefix={arXiv},
      primaryClass={cs.CV},
      url={https://arxiv.org/abs/2605.22311}, 
}

@inproceedings{photomaker,
  title        = {PhotoMaker: Customizing Realistic Human Photos via Stacked ID Embedding},
  author       = {Li, Zhen and Cao, Mingdeng and Wang, Xintao and Qi, Zhongang and Cheng, Ming-Ming and Shan, Ying},
  booktitle    = {IEEE/CVF Conference on Computer Vision and Pattern Recognition (CVPR)},
  year         = {2024}
}

@article{instantid,
  title        = {InstantID: Zero-shot Identity-Preserving Generation in Seconds},
  author       = {Wang, Qixun and Bai, Xu and Wang, Haofan and Qin, Zekui and Chen, Anthony and Li, Huaxia and Tang, Xu and Hu, Yao},
  journal      = {arXiv preprint arXiv:2401.07519},
  year         = {2024}
}

@inproceedings{ablatingconcepts,
  title        = {Ablating Concepts in Text-to-Image Diffusion Models},
  author       = {Kumari, Nupur and Zhang, Bingliang and Wang, Sheng-Yu and Shechtman, Eli and Zhang, Richard and Zhu, Jun-Yan},
  booktitle    = {IEEE/CVF International Conference on Computer Vision (ICCV)},
  pages        = {22691--22702},
  year         = {2023}
}

@inproceedings{uce,
  title        = {Unified Concept Editing in Diffusion Models},
  author       = {Gandikota, Rohit and Orgad, Hadas and Belinkov, Yonatan and Materzy{\'n}ska, Joanna and Bau, David},
  booktitle    = {IEEE/CVF Winter Conference on Applications of Computer Vision (WACV)},
  year         = {2024}
}

@inproceedings{selectiveamnesia,
  title        = {Selective Amnesia: A Continual Learning Approach to Forgetting in Deep Generative Models},
  author       = {Heng, Alvin and Soh, Harold},
  booktitle    = {Advances in Neural Information Processing Systems (NeurIPS)},
  year         = {2023}
}

@inproceedings{rombach2022ldm,
  title        = {High-Resolution Image Synthesis with Latent Diffusion Models},
  author       = {Rombach, Robin and Blattmann, Andreas and Lorenz, Dominik and Esser, Patrick and Ommer, Bj{\"o}rn},
  booktitle    = {IEEE/CVF Conference on Computer Vision and Pattern Recognition (CVPR)},
  pages        = {10684--10695},
  year         = {2022}
}

@misc{genmu2026,
  author       = {{GenMu 2026 Organisers}},
  title        = {{GenMu 2026}: Generative Model Unlearning Challenge},
  year         = {2026},
  howpublished = {Challenge at the ECCV 2026 Workshop on Unlearning
                  and Model Editing (U\&ME)},
  note         = {\url{https://iab-iitj.github.io/genmu2026/index.html}},
}

@InProceedings{antidream,
    author    = {Van Le, Thanh and Phung, Hao and Nguyen, Thuan Hoang and Dao, Quan and Tran, Ngoc N. and Tran, Anh},
    title     = {Anti-DreamBooth: Protecting Users from Personalized Text-to-image Synthesis},
    booktitle = {Proceedings of the IEEE/CVF International Conference on Computer Vision (ICCV)},
    month     = {October},
    year      = {2023},
    pages     = {2116-2127}
}

@INPROCEEDINGS{cao,
  author={Cao, Yinzhi and Yang, Junfeng},
  booktitle={2015 IEEE Symposium on Security and Privacy}, 
  title={Towards Making Systems Forget with Machine Unlearning}, 
  year={2015},
  volume={},
  number={},
  pages={463-480},
  doi={10.1109/SP.2015.35}}

@inproceedings{bourtoule,
  title={Machine unlearning},
  author={Bourtoule, Lucas and Chandrasekaran, Varun and Choquette-Choo, Christopher A and Jia, Hengrui and Travers, Adelin and Zhang, Baiwu and Lie, David and Papernot, Nicolas},
  booktitle={2021 IEEE symposium on security and privacy (SP)},
  pages={141--159},
  year={2021},
  organization={IEEE}
}

@article{nguyen,
author = {Nguyen, Thanh Tam and Huynh, Thanh Trung and Ren, Zhao and Nguyen, Phi Le and Liew, Alan Wee-Chung and Yin, Hongzhi and Nguyen, Quoc Viet Hung},
title = {A Survey of Machine Unlearning},
year = {2025},
issue_date = {October 2025},
publisher = {Association for Computing Machinery},
address = {New York, NY, USA},
volume = {16},
number = {5},
issn = {2157-6904},
url = {https://doi.org/10.1145/3749987},
doi = {10.1145/3749987},
journal = {ACM Trans. Intell. Syst. Technol.},
month = sep,
articleno = {108},
numpages = {46}
}

@INPROCEEDINGS{serfiq,
  author={Terhörst, Philipp and Kolf, Jan Niklas and Damer, Naser and Kirchbuchner, Florian and Kuijper, Arjan},
  booktitle={2020 IEEE/CVF Conference on Computer Vision and Pattern Recognition (CVPR)}, 
  title={SER-FIQ: Unsupervised Estimation of Face Image Quality Based on Stochastic Embedding Robustness}, 
  year={2020},
  volume={},
  number={},
  pages={5650-5659},
  doi={10.1109/CVPR42600.2020.00569}}

@INPROCEEDINGS{grafiqs,
  author={Kolf, Jan Niklas and Damer, Naser and Boutros, Fadi},
  booktitle={2024 IEEE/CVF Conference on Computer Vision and Pattern Recognition Workshops (CVPRW)}, 
  title={GraFIQs: Face Image Quality Assessment Using Gradient Magnitudes}, 
  year={2024},
  volume={},
  number={},
  pages={1490-1499},
  doi={10.1109/CVPRW63382.2024.00156}}

@inproceedings{circumventing,
 author = {Pham, Minh and Marshall, Kelly and Cohen, Niv and Mittal, Govind and Hegde, Chinmay},
 booktitle = {International Conference on Learning Representations},
 editor = {B. Kim and Y. Yue and S. Chaudhuri and K. Fragkiadaki and M. Khan and Y. Sun},
 pages = {1156--1181},
 title = {Circumventing Concept Erasure Methods For Text-To-Image Generative Models},
 url = {https://proceedings.iclr.cc/paper_files/paper/2024/file/048a5343fc1389f541bea87a4612c585-Paper-Conference.pdf},
 volume = {2024},
 year = {2024}
}

@misc{statistical,
      title={SMI: Statistical Membership Inference for Reliable Unlearned Model Auditing}, 
      author={Jialong Sun and Zeming Wei and Jiaxuan Zou and Jiacheng Gong and Jie Fu and Chengyang Dong and Heng Xu and Jialong Li and Bo Liu},
      year={2026},
      eprint={2602.01150},
      archivePrefix={arXiv},
      primaryClass={cs.LG},
      url={https://arxiv.org/abs/2602.01150}, 
}

@InProceedings{verification,
  title = 	 {Verification of Machine Unlearning is Fragile},
  author =       {Zhang, Binchi and Chen, Zihan and Shen, Cong and Li, Jundong},
  booktitle = 	 {Proceedings of the 41st International Conference on Machine Learning},
  pages = 	 {58717--58738},
  year = 	 {2024},
  editor = 	 {Salakhutdinov, Ruslan and Kolter, Zico and Heller, Katherine and Weller, Adrian and Oliver, Nuria and Scarlett, Jonathan and Berkenkamp, Felix},
  volume = 	 {235},
  series = 	 {Proceedings of Machine Learning Research},
  month = 	 {21--27 Jul},
  publisher =    {PMLR},
  url = 	 {https://proceedings.mlr.press/v235/zhang24h.html},
}
\end{document}